\documentclass[times, review, 10pt]{elsarticle}

\usepackage{amssymb}
\usepackage{amsmath}

\usepackage{url} 
\usepackage{float}
\usepackage{xcolor}
\usepackage{multirow}
\usepackage{booktabs}
\usepackage{array}   
\usepackage{subfigure}
\usepackage{colortbl}
\usepackage{graphicx}
\usepackage{changepage} 
\usepackage{makecell}

\definecolor{mygray}{gray}{.9}
\definecolor{green}{rgb}{0.0, 0.39, 0.0}
\journal{Pattern Recognition}

\begin{document}

\begin{frontmatter}



\title{Deformable Convolution Module with Globally Learned Relative Offsets for Fundus Vessel Segmentation} 

 \author[label1]{Lexuan Zhu}
\affiliation[label1]{organization={New York University},
            city={New York City},
             postcode={10003},
            state={New York State},
             country={United States}}

 \author[label2]{Yuxuan Li}
\affiliation[label2]{organization={The University of Sydney},
            city={Sydney},
             postcode={NSW 2006},
            state={New South Wales},
             country={Australia}}
 \author[label3]{Yuning Ren\corref{cor1}}
\affiliation[label3]{organization={China University of Political Science and Law},
	postcode={100088},
	state={Beijing},
	country={China}}
\cortext[cor1]{Corresponding author: Y. R. yuning.ren.cupl.cn@gmail.com}

%

\begin{abstract}
Deformable convolution can adaptively change the shape of convolution kernel by learning offsets to deal with complex shape features. We propose a novel plug-and-play deformable convolutional module that uses attention and feedforward networks to learn offsets, so that the deformable patterns can capture long-distance global features.
Compared with previously existing deformable convolutions, the proposed module learns the sub-pixel displacement field and adaptively warps the feature maps across all channels (rather than directly deforms the convolution kernel), which is equivalent to a relative deformation of the kernel’s sampling grids, achieving global feature deformation and the decoupling of (kernel size)–(learning network).
Considering that the fundus blood vessels have globally self-similar complex edges, we design a deep learning model for fundus blood vessel segmentation, GDCUnet, based on the proposed convolutional module. Empirical evaluations under the same configuration and unified framework show that GDCUnet has achieved state-of-the-art performance on public datasets.
Further ablation experiments demonstrated that the proposed deformable convolutional module could more significantly learn the complex features of fundus blood vessels, enhancing the model's representation and generalization capabilities.
The proposed module is similar to the interface of conventional convolution, we suggest applying it to more machine vision tasks with complex global self-similar features.
Our source code will be available at \url{https://github.com/LexyZhu/GDCUnet.git}
\end{abstract}

\begin{graphicalabstract}
\includegraphics[width=1.0\textwidth]{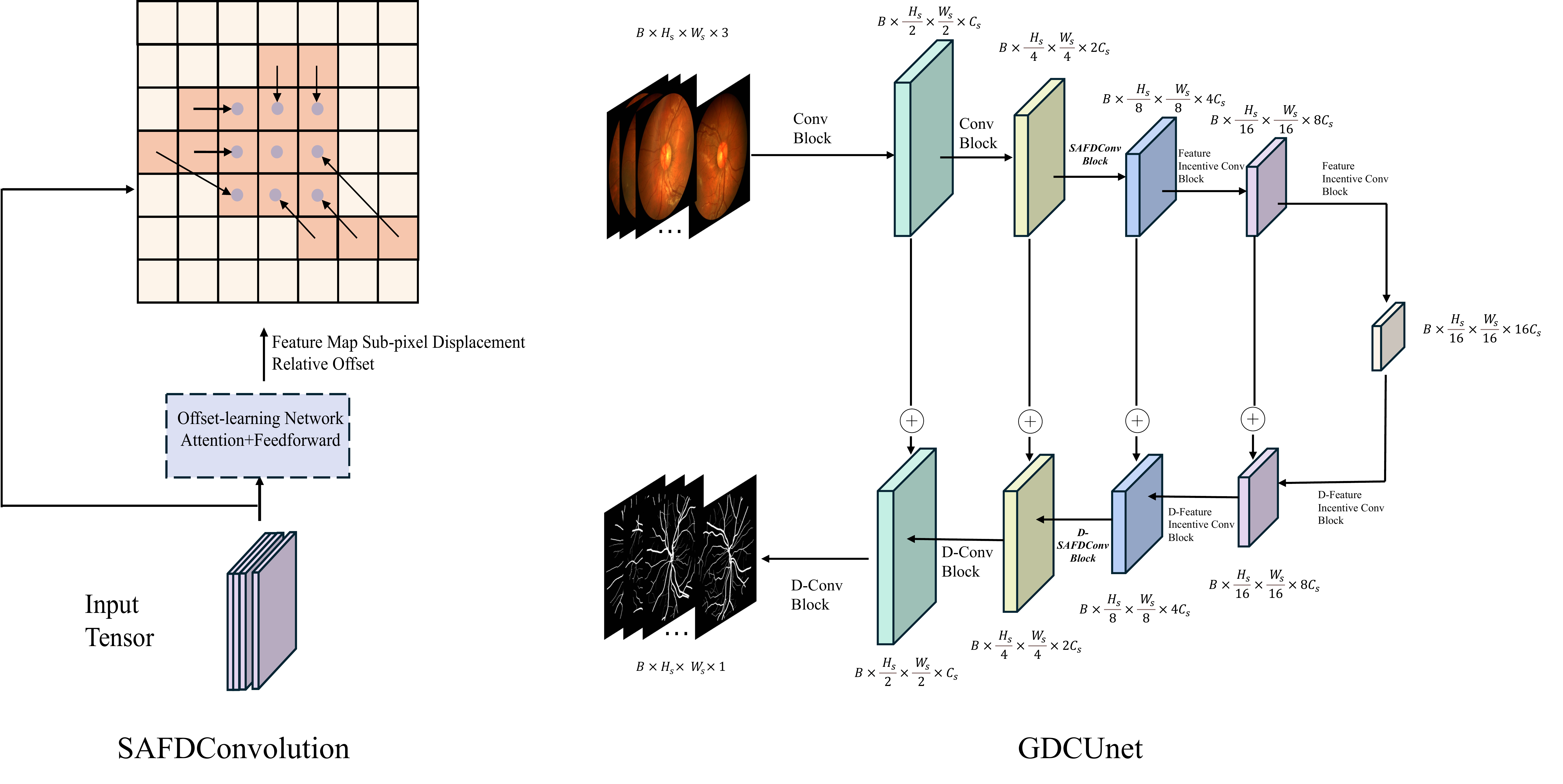}
\includegraphics[width=1.2\textwidth]{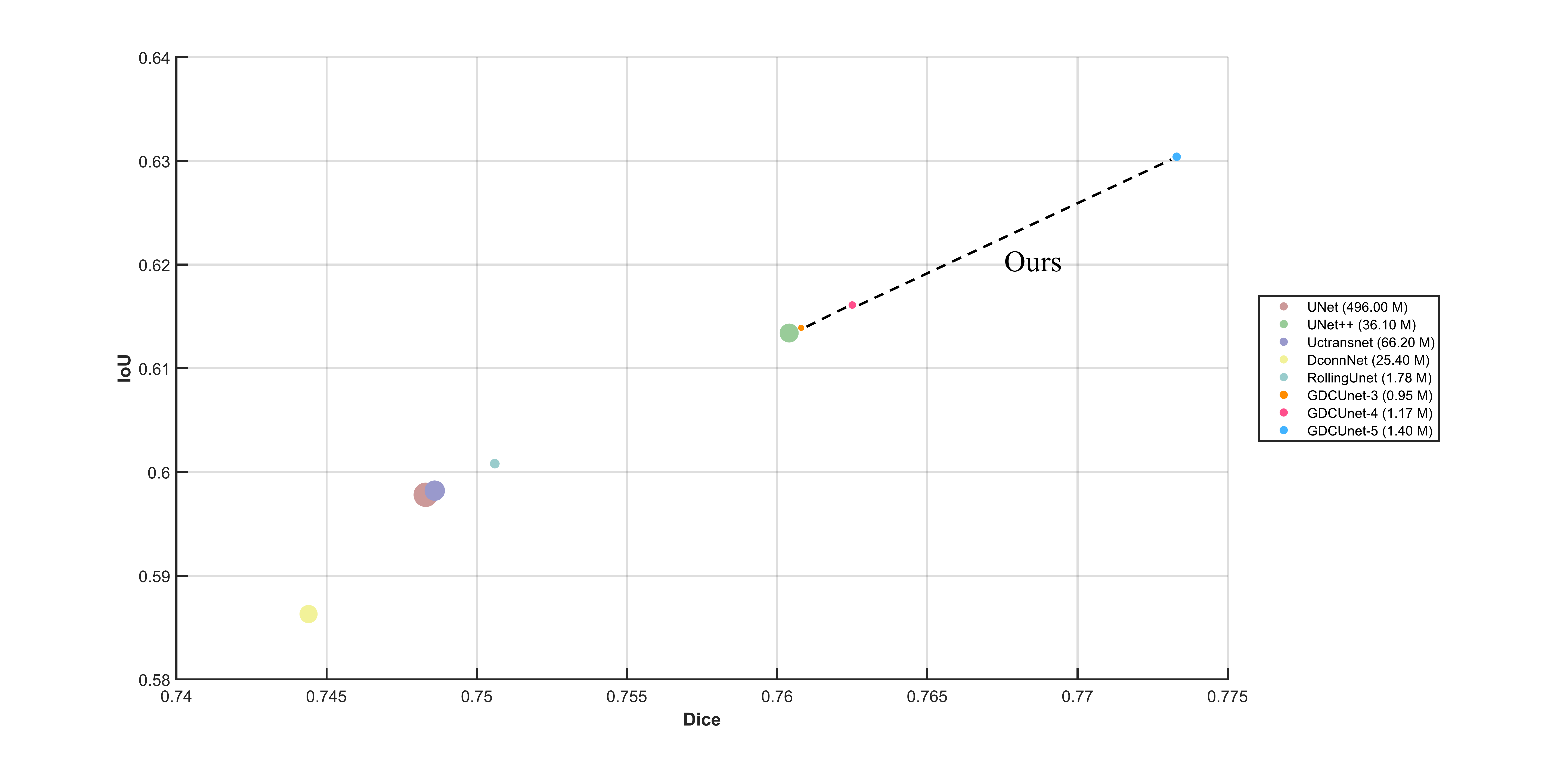}
\end{graphicalabstract}

\begin{highlights}
\item We propose a plug-and-play deformable convolution with globally learned relative offsets, a module well suited to handle complex zigzag edge features. Compared with other existing deformable convolution, the proposed convolution has the following different characteristics:

 \begin{itemize}
	\item Offsets are learned using multi-head attention and feedforward networks, which can effectively capture global semantic dependencies.
	\item The offset acts on the convolutional feature map, forming a sub-pixel-level deformation field, thereby causing the convolutional kernel to undergo "relative deformation".
	\item The module can decouple the size of the convolution kernel and the offset learning network, and can adapt to conventional convolution with any parameter.
\end{itemize}

\item We add the proposed convolutional module to Unet and propose a deep learning model for fundus vascular segmentation, called GDCUnet. 
\item We construct a unified comparison framework with the same data processing methods and loss functions and other configurations for the mainstream existing state-of-the-art medical image segmentation models in our project. Experiments on the public dataset have verified that GDCUnet has reached state-of-the-art performance.
\end{highlights}

\begin{keyword}
deformable convolution \sep plug-and-play module \sep medical image segmentation \sep attention \sep relative deformation.
\end{keyword}

\end{frontmatter}



\section{Introduction}
\label{int}
Medical image segmentation technology extracts the region of interest as a binary mask and has important applications in analysis and diagnosis \cite{MS}.
This task can be carried out manually by professional medical workers. However, accurate computer-aided segmentation can enhance efficiency and objectivity \cite{Background}.
Unet is a U-shaped structure based on convolution and residual feature reuse, and has become an important solution for tasks such as biological image segmentation \cite{Unet}. Some improved versions of Unet can achieve better performance in certain tasks \cite{FDUnet,Unetadd}.
The global feature learning ability of convolution has limitations. Vision Transformer (ViT) uses the attention mechanism for visual modalities to enhance long-distance feature learning \cite{ViT}. Medical images often have self-similar global features, so ViT-like structures have outstanding analytical and understanding capabilities \cite{MedT}.
However, the Transformer structure is slightly weak in local feature representation and multi-scale feature fusion.
In recent years, the improvement direction of machine vision models has been the fusion understanding of local and global features \cite{LoGoNet}. Medical image segmentation tasks also require the integration of global and local features to achieve better performance \cite{Rollingunet}.

\begin{figure}[H]
	\centering
	\includegraphics[width=0.6\textwidth]{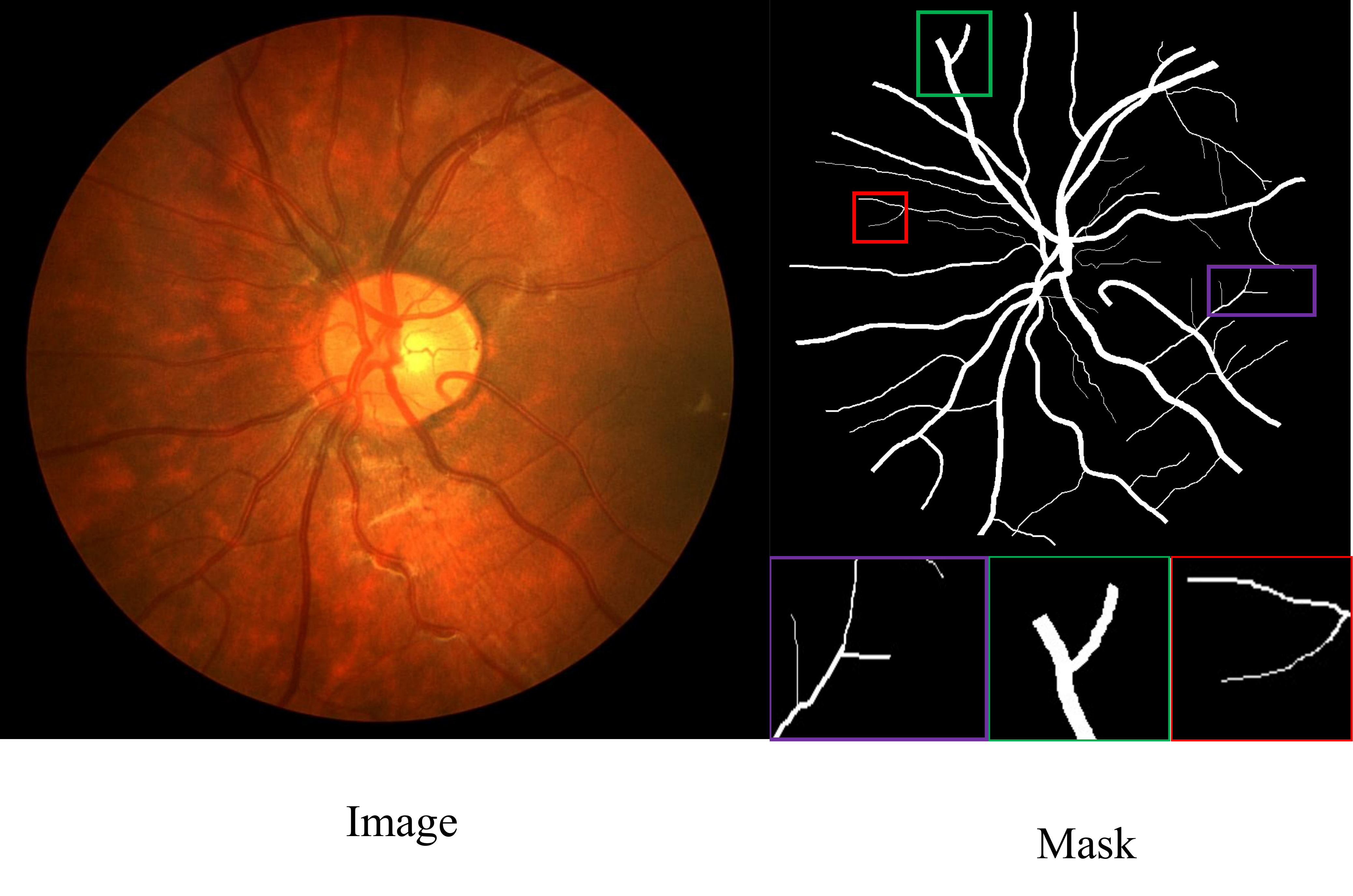}
	\caption{Local and global features of fundus vessels.}
	\label{fig:1}
\end{figure}

As shown in Figure~\ref{fig:1}, blood vessels have subtle local features and at the same time are globally self-similar. This paper focuses on \textbf{how to enhance the ability of deep learning models to represent locally complex but globally self-similar vascular features}, so as to improve segmentation performance.
We consider the learning of deformable convolutional offsets as a solution to this problem.

Deformable convolution is based on the offset learned by a network to adaptively change the shape of the convolution kernel, enabling the convolution to focus on the features of objects with complex shapes \cite{DCN}.
Deformable convolution is highly suitable for handling the complex local edges of medical images \cite{DCMI}.
However, traditional deformable convolution relies on another convolution to learn the offset, which limits its representation of global self-similar feature shape.
In this paper, we use \textbf{S}patial \textbf{A}ttention and \textbf{F}eedforward network to learn offsets and propose a novel \textbf{D}eformable \textbf{Convolution}, which we name \textbf{SAFDConvolution}. SAFDConvolution is plug-and-play and has a similar interface to regular convolutions, making it very convenient to replace regular convolutions in convolutional neural networks. Based on this, we design a fundus vascular image segmentation model similar to Unet using conventional convolution and the proposed convolution, the \textbf{G}lobally \textbf{D}eformable \textbf{C}onvolution \textbf{Unet} (\textbf{GDCUnet}).

Considering that the current medical image segmentation models lack a unified comparison framework, which limits the persuasiveness of comparison experiments, we construct a unified configuration framework and uniformly configured some of the existing state-of-the-art models. Experiments on the public dataset show that GDCUnet achieves best performance with the minimum number of parameters.
Our main contributions are summarized in the \textbf{Highlights} above, the rest of this paper is organized as follows: 
Section \ref{rel} reviews the works related to local global models and deformable convolutions. In Section \ref{met}, we propose the SAFDConvolution and GDCUnet. Section \ref{exp} presents the main results of the experiment. Section \ref{abl} demonstrates the feature learning ability and generalization ability of the proposed method through ablation experiments and Section \ref{con} concludes the entire paper.

\section{Related Work}
\label{rel}
\subsection{Local–Global Model for Medical Image Processing}
The global features of medical images have gradually attracted attention, and thus medical image analysis models similar to ViT have been proposed \cite{MedT}. Given that pure Transformers lack the ability to grasp local details, some works focus on using local-global multi-scale feature fusion understanding models. The processing methods include local window attention structures \cite{Swinunet} similar to Swin Transformer \cite{SwinT}, and some other works are the combination of convolution and attention \cite{SpachTransformer,Uctransnet,Age,MCTE}. Considering the computational complexity and other issues, some models abandon the traditional attention mechanism to process the long-distance semantic dependence of the convolutional feature map \cite{Rollingunet,GLFRNet}. The local-global concept is also applied to handle medical images of higher dimensions and more modalities \cite{3DLG}.

Blood vessels have more unique features that exhibit globally self-similar local complex edges, which inspires us to find the new solution using convolution and attention.

\subsection{Applications of Spatial Attention}
Many studies have been conducted on methods to enhance the spatial global feature representation ability of convolutional neural networks \cite{Nonlocal,CBAM,CAttention}.
In recent years, spatial attention models for different vision tasks have been proposed \cite{SAWUNet,LKSSAN}, some requiring the combination of channel attention to further enhance learning ability \cite{CSCANet,SLCAM}.
For continuous video frames, spatial attention can be combined with temporal attention to efficiently process temporal continuous global features \cite{ST1,ST2,ST3}.

The success of spatial attention has provided us with inspiration for processing methods of global self-similar features with complex shape.

\subsection{Convolutions with Diverse Sampling Patterns}
\paragraph{Deformable Convolution V1 \cite{DCN}}

J. Dai et al. first proposed deformable convolution, which uses another dedicated convolutional layer to learn offsets, achieving adaptive changes in the shape of convolutional sampling points and enhancing the adaptability of the convolutional layer to complex shapes.

\paragraph{Deformable Convolution V2 \cite{V2}}
V2 introduces modulation amplitude control and adopts the concept of feature mimicking to restrict deformation behavior at the feature points of interest.

\paragraph{Deformable Convolution V3 \cite{V3}}
V3 sparsifies the sampling kernel and adopts the concept of spatial aggregation, enhancing the long-distance perception ability of the deformable convolution and achieving a competitive performance with the ViT structure.

\paragraph{Deformable Convolution V4 \cite{V4}}
V4 removes the softmax normalization in the spatial aggregation of V3 to further enhance dynamics and expressiveness. The main focus of V4 is computational efficiency, reducing redundant operations and enhancing inference efficiency.
\paragraph{Others}
To enlarge the receptive field without increasing the parameter count, F. Yu et al. introduced dilation into the convolution kernel, thereby proposing dilated convolution \cite{Dilated}.
In addition, dedicated convolution for special tubular topological structures has been studied, and this convolution is called dynamic snake convolution \cite{Snake}.

However, the current concept of all deformable convolutions is the offset of each sampling grid, which can not solve the decoupling of the convolution kernel size and the offset learning network, and has limitations on the integration friendliness of global learning modules such as attention.

\section{Method}
\label{met}
\subsection{SAFDConvolution}
Suppose the convolution kernel size \(\mathbf{K_s}\) is one odd number, make radius \(r=\tfrac{\mathbf {K_s} - 1}{2}\), then the 2D conventional sampling grid \(\mathcal{R}\) with dilation rate \(\mathbf{D_s}\) is defined as:

\begin{align}
	\mathcal{R}_{\mathbf{K_s},\mathbf{D_s}}
	=\bigl\{\, (i\mathbf{D_s},\; j\mathbf{D_s})\;\bigm|\;
	i,j\in\mathbb{Z},\;-r\le i\le r,\;-r\le j\le r \bigr\}.
	\label{eq.grid_Ks_Ds}
\end{align}
For example,  
\begin{itemize}
	\item \(\mathbf{K_s}=3,\;\mathbf{D_s}=1 \;\Rightarrow\;\)  \(3\times3\) grid \(\{(-1,-1),(-1,0),(0,-1),\dots,(1,1)\}\);
	\item \(\mathbf{K_s}=3,\;\mathbf{D_s}=2 \;\Rightarrow\;\) \(5\times5\) grid \(\{(-2,-2),(-2,0),(0,-2),\dots,(2,2)\}\).
\end{itemize}

Given the input feature map \(\mathbf{x}(\mathbf{p}_0)\) with spatial coordinates of \(\mathbf{p}_0 \in\mathbb{Z}^2\), the standard convolution with dilation rate $\mathbf{D_s}$ is written as
\begin{align}
	\mathbf{y}(\mathbf{p}_0)=
	\sum_{\mathbf{p}_n \in \mathcal{R}_{\mathbf{K_s},\mathbf{D_s}}}
	\mathbf{w}(\mathbf{p}_n)\;
	\mathbf{x}\bigl(\mathbf{p}_0+\mathbf{p}_n\bigr),
	\label{eq.standard_dilated_conv}
\end{align}
where \(\mathbf{w}(\mathbf{p}_n)\) represents the convolution kernel weights. To enable the convolution kernel to adaptively undergo geometric deformation, the existing deformable convolutions introduce an a learnable offset \(\{\Delta\mathbf{p}_n\}\):
\begin{equation}
	\mathbf{y}(\mathbf{p}_0)=
	\sum_{\mathbf{p}_n \in \mathcal{R}_{\mathbf{K_s},\mathbf{D_s}}}
	\mathbf{w}(\mathbf{p}_n)\;
	\mathbf{x}\bigl(\mathbf{p}_0+\mathbf{p}_n+\Delta\mathbf{p}_n\bigr).
	\label{eq.deformable_dilated_conv}
\end{equation}
Instead of learning a set of discrete offsets
\(\{\Delta\mathbf p_n\}_{n=1}^{K_s^{2}}\) for every kernel location \(\mathbf{p_n}\) as in
Eq.~\eqref{eq.deformable_dilated_conv},
we learn a \emph{continuous} \emph{sub-pixel displacement field:}
\begin{align}
	\Delta\mathbf p:\ \mathbb Z^{2}\longrightarrow\mathbb R^{2},
	\qquad
	\mathbf p\mapsto\Delta\mathbf p(\mathbf p)
\end{align}
whose value is shared by all channels at the same spatial coordinate. The feature map is first warped by this field:
\begin{align}
	\tilde{\mathbf x}(\mathbf p)
	\;=\;
	\mathbf x(\mathbf p+\Delta\mathbf p(\mathbf p)),
	\label{eq.warping}
\end{align}
and the regular convolution is then applied to~$\tilde{\mathbf x}$:
\begin{align}
	\mathbf y(\mathbf p_0)
	&=
	\sum_{\mathbf p_n\in\mathcal R_{K_s,D_s}}
	\mathbf w(\mathbf p_n)\;
	\tilde{\mathbf x}(\mathbf p_0+\mathbf p_n)
	\notag\\
	&=
	\sum_{\mathbf p_n\in\mathcal R_{K_s,D_s}}
	\mathbf w(\mathbf p_n)\;
	\mathbf x(\mathbf p_0+\mathbf p_n
	+\Delta\mathbf p(\mathbf p_0+\mathbf p_n)).
	\label{eq.relative_deform_conv}
\end{align}

\begin{figure}[H]
	\centering
	\includegraphics[width=1.0\textwidth]{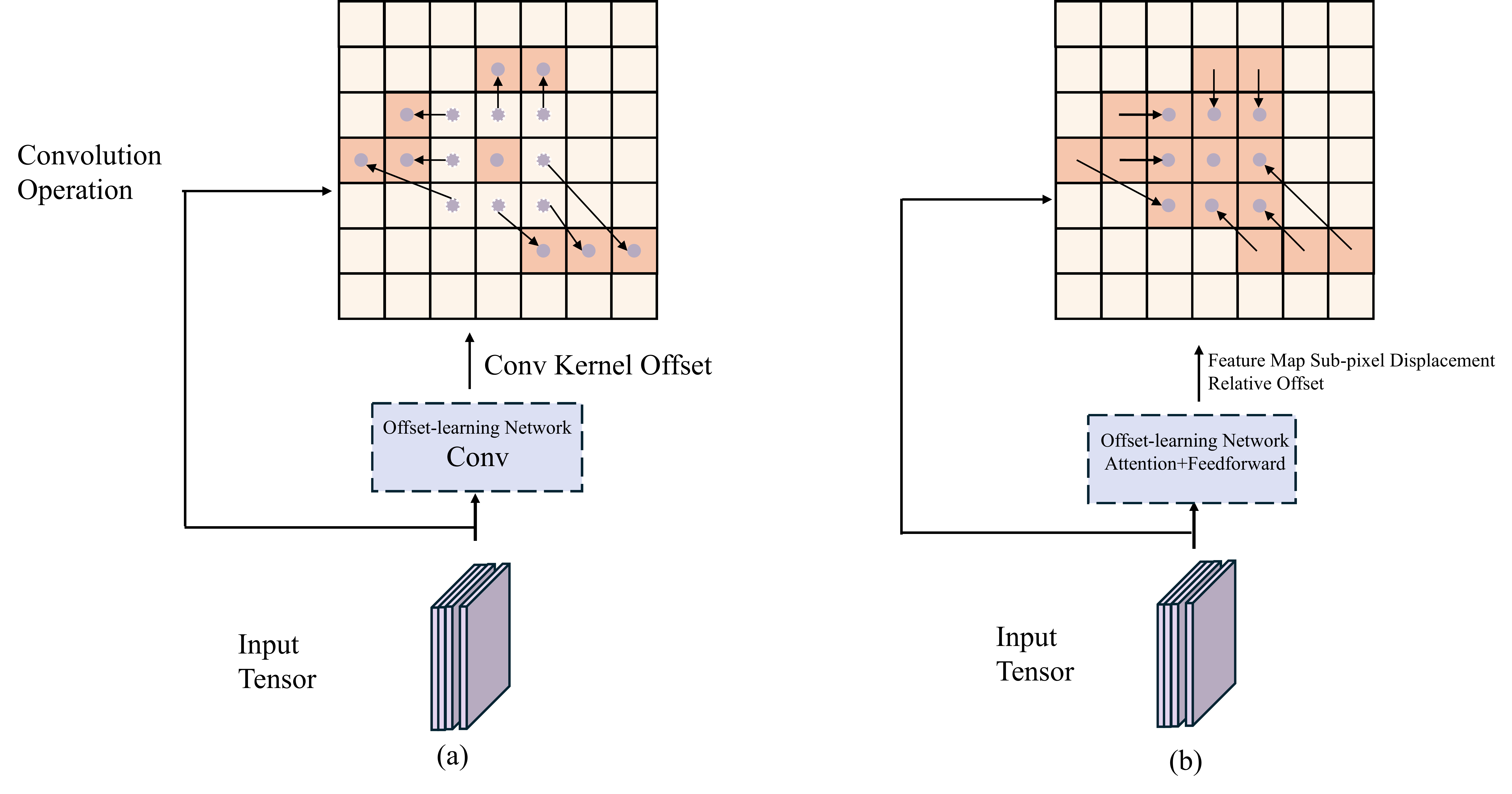}
	\caption{Conceptual differences between SAFDConvolution and existing deformable convolutions, (a) Existing deformable convolutions, (b) Our SAFDConvolution. SAFDConvolution does not directly change the shape of the convolution kernel but deforms the feature map to achieve relative deformation of the convolution kernel. The displacement field of the feature map is shared for all channels. The learning of relative offsets achieves the capture of globally self-similar complex local details through attention and feedforward networks (like the ViT).}
	\label{fig:SAFDConvolution}
\end{figure}

To obtain the offset \(\Delta\mathbf p(\mathbf p)\) by the attention, for the input tensor \(X\in\mathbb{R}^{B\times H\times W\times C}\):
\begin{align}
	X_u &= \operatorname{reshape}(X,\;B,HW,C) \in \mathbb{R}^{B \times N \times C}
\end{align}
where \(N=HW\), then we introduce an optional hyperparameter \(\mathcal{E}\) to enhance the feature dimension. Based on this hyperparameter, the feature dimension is embedded through a linear layer:
\begin{align}
	X_e &= X_u\,W^{E} &&\in \mathbb{R}^{B \times N \times \mathcal{E}C}
\end{align}
where \(W^{E}\) is the weight matrix of the embedding layer, mark \(\mathcal{E}C\) as \(D\).
The attention mechanism \(X_e\) into three tensors, Query, Key and Value, through a linear transformation:
\begin{align}
	Q &= X_e\,W^{Q},\quad K = X_e\,W^{K},\quad V = X_e\,W^{V} &&\in \mathbb{R}^{B \times N \times D} 
\end{align}
Learning from ViT, the 3 tensors \(Q\), \(K\) and \(V\) are chunked into \(h\) heads:

\begin{align}
	Q_{h_{i}} &= \operatorname{chunk_{h_{i}}}(Q,\;B,N,d_h)              &&\in \mathbb{R}^{B  \times N \times d_h} \\[2pt]
	K_{h_{i}} &= \operatorname{chunk_{h_{i}}}(K,\;B,N,d_h)              &&\in \mathbb{R}^{B \times N \times d_h} \\[2pt]
	V_{h_{i}} &= \operatorname{chunk_{h_{i}}}(V,\;B,N,d_h)              &&\in \mathbb{R}^{B  \times N \times d_h} 
\end{align}
where \(d_h=\tfrac{D}{h}\), \(h_{i} \in {1,2,\dots,h}\). Then perform the regular attention multiplication:

\begin{align}
	\mathcal{A}_{h_{i}} &= \operatorname{SoftMax}(\tfrac{Q_{h_{i}} K_{h_{i}}^{\!\top}}{\sqrt{d_h}})V_{h_{i}} &&\in \mathbb{R}^{B \times N \times d_h} 
\end{align}
where \(\mathcal{A}_{h_{i}}\) represents the output at the \(h_{i}\)-th head, the output of all heads is concatenated to obtain the output of attention \(\mathcal{A}\):
\begin{align}
	\mathcal{A} =  \text{Concat}[\mathcal{A}_{1}, \cdots, \mathcal{A}_{h}]W^O &&\in \mathbb{R}^{B  \times N \times D} 
\end{align}
where \(W^O\) is another linear weight matrix. We perform the residual operation before inputting the feedforward network:
\begin{align}
	X_A =  \mathcal{A} + X_e &&\in \mathbb{R}^{B  \times N \times D} 
\end{align}
Then we provide another optional hyperparameter \(D_{hidden}\) as the hidden dimension of feedforward, let \(X_A\) pass through two linear layers:
\begin{align}
	X_{hidden} =  X_A W_{hidden}^{(1)} &&\in \mathbb{R}^{B  \times N \times D_{hidden}} 
\end{align}
\begin{align}
	X_{F} =  X_{hidden} W^{(2)} &&\in \mathbb{R}^{B  \times N \times D} 
\end{align}
where \(W_{hidden}^{(1)} \in \mathbb{R}^{B  \times D \times D_{hidden}} \), \(W^{(2)} \in \mathbb{R}^{B  \times D_{hidden} \times D} \). 
After that, we perform another residual operation and reverse embedding:
\begin{align}
	\hat{X} =  	(X_{F} + X_{A})W^{E}_{reverse} &&\in \mathbb{R}^{B  \times N \times C} 
\end{align}
where \(W^{E}_{reverse} \in \mathbb{R}^{B  \times D \times C} \).  \(\hat{X}\) needs to be reshaped back into the form of a convolutional feature map:
\begin{align}
	X_{offset} &= \operatorname{reshape}(\hat{X},\;B,H,W,C) \in \mathbb{R}^{B \times H \times W \times C}
\end{align}
The number of channels \(C\) is usually divisible by \(2\), we denote that \(M_c=C/2\) and split the last dimension:

\begin{align}
	\tilde{X}_{offset} &= \operatorname{reshape}(X_{offset},\;B,H,W,M_c,2) \in \mathbb{R}^{B \times H \times W \times M_c \times 2}
\end{align}
Then we take the average of \(M_c\) tensors to obtain the final \(\Delta\mathbf p(\mathbf p_0+\mathbf p_n)\), which is shared by all channels:

\begin{align}
	\Delta\mathbf p(\mathbf p_0+\mathbf p_n) &\xleftarrow{} 
	\frac{1}{M_c}\sum_{m_c=1}^{M_c}
	\tilde{X}_{offset}
	&&\in\mathbb{R}^{2} \text{   for each point  in  } B \times H \times W;
\end{align}
At last, sub‑pixel displacement field is added to every sampling point \(\mathbf p_0+\mathbf p_n\) in Eq.~\eqref{eq.relative_deform_conv}. Figure~\ref{fig:SAFDConvolution} illustrates the differences between SAFDConvolution and existing deformable convolutions, this structure achieves decoupling of (kernel size) - (learning network), allowing for the diversity of the offset-learning network. The sharing of relative offsets for all channels realizes the lightweight of the number of parameters.

We summarize the selectable hyperparameters of SAFDConvolution in Table \ref{tab:hyperparameters} and provide some hyperparameter settings for subsequent experiments (in Section \ref{results}).

	\begin{table}[H]
	\centering
	
	\caption{Hyperparameter Settings of SAFDConvolution for Experiments. \(\mathbf{K_s}\)-Kernel Size, \(\mathbf{D_s}\)-Dilation Rate, \(\mathcal{E}\)-Multiple of Embedding, \(h\)-Number of Attention Heads, \(D_{hidden}\)-Hidden Dimension of Feedforward. }
	\begin{tabular}{l c c c c c c}
		\toprule
		& Setting 1&    Setting 2& Setting 3 & Setting 4 & Setting 5 & Setting 6   \\
		\midrule 
		\(\mathbf{K_s}\)  &5 &7 & 3 & 5& 5& 5    \\
		\(\mathbf{D_s}\) &1&1 & 2& 1& 1& 1 \\
		\(\mathcal{E}\) &1&1 &1 & 1& 2& 4  \\
		\(h\) & 4& 4& 4& 4& 4& 4 \\
		\(D_{hidden}\)&32&32 &32 & 64& 64& 64 \\

		\bottomrule
	\end{tabular}

	\label{tab:hyperparameters}
\end{table}

\subsection{GDCUnet}

\paragraph{Encoding Stage}
During the encoding stage, the Conv Block contains two \(3 \times 3\) conventional convolutions. SAFDConv Block uses two consecutive proposed SAFDConvolutions, where the hyperparameters can be set according to Table~\ref{tab:hyperparameters}. Further, the Feature Incentive Block uses one \(7 \times 7\) conventional convolution.

\paragraph{Decoding Stage}

The structure block used in the decoding stage (D-Block) is the same as modules adopted in the encoding stage.

\paragraph{Overall Pipeline}

\begin{figure}[H]
	\centering
	\includegraphics[width=1.0\textwidth]{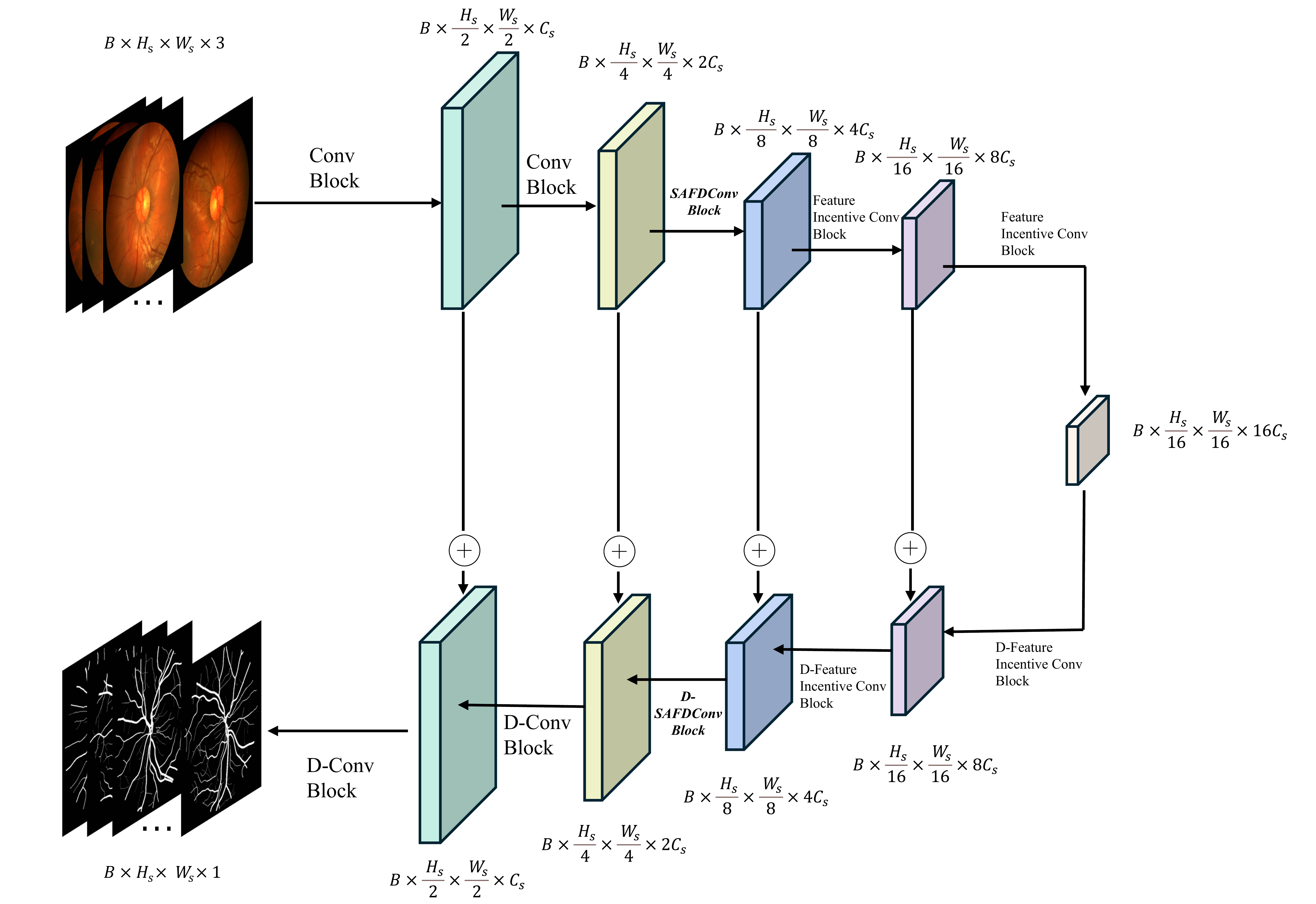}
	\caption{Overall pipeline of GDCUnet, where \(C_s\) is \(16\). Our network adopts a structure similar to Unet. In the encoding stage, Max-pooling is used to reduce the size of the feature map, and in the decoding stage, bilinear interpolation is used to restore the size of the feature map. We use tensor addition to achieve feature fusion of the feature maps from the previous stage. The modules using proposed SAFDConvolution are \textit{\textbf{italicized bold.}}. To the best of our knowledge, the proposed deformable convolution is more suitable in the intermediate stage of feature extraction and decoding.}
	\label{fig:GDCUnet}
\end{figure}

Suppose the shape of input tensor is \(B \times H_s \times W_s \times 3\), the pipeline of GDCUnet is illustrated in Figure~\ref{fig:GDCUnet}, the final output is a 1-channel segmentation mask \(B \times H_s \times W_s \times 1\).

\section{Experiments}
\label{exp}
\subsection{Unified Benchmarks and Configurations}

To the best of our knowledge, unlike multi-dimensional time series analysis (e.g.\cite{liu2024itransformer}), the current medical image segmentation models do lack a unified framework when comparing representation and generalization capabilities, resulting in differences in configurations such as regularization methods, loss functions, and learning rates. This limits the full explanation of the algorithm's performance. We have established a unified benchmark and made compatibility processing for some classic models and existing state-of-the-art models. For more details, please visit our project website. We hope this will be helpful for the comparative experiments of other future works.

\subsection{Implementation Details}
Our experiments are built on CHASEDB1 \cite{Dataset}, a public retinal blood vessel reference dataset.
The dataset is divided into a training set and a test set in a ratio of \(0.86:0.14\).
The experimental configuration summary is presented in Table \ref{tab:exp}. All experiments are conducted using the Pytorch framework \cite{Pytorch}, and the learning rate is uniformly optimized using cosine annealing. We evaluate the model on the test set after each epoch and report the highest score achieved.

	\begin{table}[H]
	
	\centering
	
	\caption{Configuration Settings for Experiments}
	\begin{tabular}{l c}
		\toprule
		Settings & Details            \\
		\midrule 
		Batch Size & 4           \\
		Initial Learning Rate & \(1 \times 10^{-4}\)   \\
		Minimum Learning Rate & \(1 \times 10^{-5}\)   \\
		Optimizer & Adam \cite{Adam} (Default \(\beta_1 = 0.0\), \(\beta_2 = 0.99\))   \\
		Input Resolution & \(256 \times 256\)   \\
		Output Resolution & \(256 \times 256\)   \\
		Total Epochs & 4,000\\
		GPU &  Nvidia Tesla V100 32G\\
		
		\bottomrule
	\end{tabular}

		\label{tab:exp}
\end{table}

The loss function BCEDiceLoss \(\mathcal{L}\) adopted in this paper is a mixture of cross-entropy and DiceLoss \cite{Loss},
given the network output logits \(p\) and the binary target \(y\in\{0,1\}\), the loss consists of two components:

\begin{align}
	\mathcal{L}_{\mathrm{BCE}}
	&= -\frac{1}{N}\sum_{i=1}^{N}
[
	y_i\,\log \sigma(p_i) +
	(1-y_i)\,\log(1-\sigma(p_i))
], \\
	\mathcal{L}_{\mathrm{Dice}}
	&= 1-\frac{2\sum_{i}\sigma(p_i)\,y_i+\varepsilon}
	{\sum_{i}\sigma(p_i)+\sum_{i}y_i+\varepsilon},
	\qquad \varepsilon = 10^{-5}, \\
	\mathcal{L}
	&= 0.5\,\mathcal{L}_{\mathrm{BCE}} + \mathcal{L}_{\mathrm{Dice}},
\end{align}
where \(\sigma(p)=\tfrac{1}{1+e^{-p}}\), \(i \in {0,1,2,\dots,N}\), \(N = H_s \times W_s\) is the sum of all pixels.

\subsection{Metrics}

We adopt recognized evaluation metrics, which are summarized in Table \ref{tab:seg_metrics}, where \(P\) denotes predicted foreground, \(G\) denotes ground‑truth foreground, \(TP\), \(FP\), \(TN\), \(FN\) denote true/false positives/negatives.
\begin{table}[H]
	\centering
	\caption{Summary of evaluation metrics, where IoU and Dice (F1 score) are the overall main indicators for evaluating the segmentation performance.}
	\label{tab:seg_metrics}
	
	\renewcommand\arraystretch{1.2}
	\setlength\tabcolsep{4pt}
	\begin{tabular}{
			>{\centering\arraybackslash}p{0.05\linewidth}
			>{\centering\arraybackslash}p{0.4\linewidth}
			>{\centering\arraybackslash}p{0.555\linewidth}}
		\toprule
		\textbf{Metric} & \textbf{Formula} & \textbf{Interpretation} \\
		\midrule
		IoU   & \( \frac{|P \cap G|}{|P \cup G|} \) & Overlap between prediction and ground truth. \\
		Dice  & \( \frac{2|P \cap G|}{|P| + |G|} \) & Harmonic mean of precision and recall. \\
		HD    & \(\begin{aligned}[t]
			\tiny
			max\{
			&\sup_{p\in P}\inf_{g\in G} d(p,g),
			\sup_{g\in G}\inf_{p\in P} d(p,g)\}
		\end{aligned}\) &
		Largest boundary error; lower is better. \\
		HD\(_{95}\) & 95\,\% quantile of HD & Less sensitive to outliers than HD. \\
		Recall       & \( \frac{TP}{TP + FN} \) & Ability to find all positives. \\
		Specificity  & \( \frac{TN}{TN + FP} \) & Ability to reject negatives. \\
		Precision    & \( \frac{TP}{TP + FP} \) & Proportion of predicted positives that are true. \\
		\bottomrule
	\end{tabular}
\end{table}

\subsection{Main Results}
\label{results}

Table~\ref{tab:results} illustrates the quantitative assessment of the segmentation performance of our GDCUNet, with various variant \textit{\textbf{Settings}} corresponding to Table~\ref{tab:hyperparameters}. The existing state-of-the-art methods compared include the improvement of model representation ability and the modeling of medical image segmentation prior knowledge (mainly considering the topological prior of edges). We label metrics that clearly highlight excellence in green.

\begin{table}[H]
	\centering
	
	\tiny
	\caption{Quantitative Comparison with State-of-the-art Methods.}\label{tab:results}
	{
		\setlength\tabcolsep{4pt}
		\renewcommand\arraystretch{1.5}
		\begin{tabular}{c|c|c|c|c|c|c|c|c|c}
			\hline

			& Model(With & & & & & & Re- &Speci- &Pre-\\
			
			\multirow{-2}{*}{Type}& \textbf{\#Parameters})  &\multirow{-2}{*}{Venue} & \multirow{-2}{*}{IoU}  & \multirow{-2}{*}{Dice}     & \multirow{-2}{*}{HD}   & \multirow{-2}{*}{HD\(_{95}\)} &call&ficity &cision \\\hline
			& Unet \textbf{\#496M} \cite{Unet}& \textit{MICCAI'2015} & 0.5978 & 0.7483& 15.62&2.000 & 0.7498& 0.9817& 0.7469\\
			&Unet++ \textbf{\#36.1M} \cite{Unetadd} & \textit{ML-CDS'2018 \footnotemark[1]} & 0.6134  & 0.7604& 15.62& 1.414&0.7620&0.9826&0.7588\\
			Model & Att-Unet \textbf{\#127M} \cite{Attentionunet} & \textit{Arxiv'2018} & 0.6039  & 0.7531 & 16.00& 2.236&0.7146&0.9868& 0.7959\\
			Enhancement & Unext \textbf{\#1.47M} \cite{Unext} & \textit{MICCAI'2022} & 0.5007  & 0.6673&17.26 & 3.000& 0.6594& 0.9772& 0.6754\\
			& Uctransnet \textbf{\# 66.2M}  \cite{Uctransnet} &  \textit{AAAI'2022}  & 0.5982 & 0.7486& 15.00 &2.000&0.7282&0.9844& 0.7702 \\
		& \makecell[l]{Rolling-Unet  \textbf{\#1.78M} (\textbf{\#7.10M})} \cite{Rollingunet} 
		& \textit{AAAI'2024} 
		& \makecell{0.6008 \\ (0.6082)} 
		& \makecell{0.7506 \\ (0.7563)} 
		& \makecell{15.00 \\ (16.00)} 
		& \makecell{1.414 \\ (1.732)} 
		& \makecell{0.7745 \\ (0.7549)} 
		& \makecell{0.9828 \\ (0.9826)} 
		& \makecell{0.7568 \\ (0.7578)} \\
		
			\hline
			Prior 
			& DconnNet \textbf{\#25.4M}  \cite{DConn} & \textit{CVPR'2023}  & 0.5863 & 0.7444 &17.76 & 3.000 &0.7929&0.9752 &0.7407\\
			Knowledge \footnotemark[2]
			& DSCNet \textbf{\#- -M} \cite{Snake}& \textit{ICCV'2023}   & 0.6073 & 0.7521& 15.00 &2.000&0.7534& 0.9832& 0.7638 \\
			\hline 
			\rowcolor{gray!20}
			& \textit{\textbf{Setting 1}}  \textbf{\#1.15M}& \textit{-----}  & 0.6160 & 0.7623& 15.56 & 2.000& 0.7495& 0.9844& 0.7757 \\
			\rowcolor{gray!20}
			& \textit{\textbf{Setting 2}}  \textbf{\#1.45M}& \textit{-----}  & 0.6201 & 0.7655& 15.00 & 1.732& 0.7511& 0.9848& 0.7805 \\
			\rowcolor{gray!20}
			\textbf{Our}& \textit{\textbf{Setting 3}}  \textbf{\#0.954M}& \textit{-----}  &  0.6139& 0.7608& 15.00& 2.000& 0.7470& 0.9844& 0.7750 \\
			\rowcolor{gray!20}
			\textbf{GDCUnet}& \textit{\textbf{Setting 4}}  \textbf{\#1.17M}& \textit{-----}  &  0.6161&0.7625 & 17.23 & 2.000 & 0.7433 & 0.9852 & 0.7827 \\
			\rowcolor{gray!20}
			& \textit{\textbf{Setting 5}}  \textbf{\#1.40M}& \textit{-----}  & \textcolor{green}{0.6304} & \textcolor{green}{0.7733}&15.36 & 1.732&0.7596& 0.9853& 0.7875  \\
			\rowcolor{gray!20}
			& \textit{\textbf{Setting 6}}  \textbf{\#2.16M}& \textit{-----}  &0.6132  & 0.7602& 15.00 & 2.000& 0.7394& 0.9852& 0.7823 \\

			\hline
			\hline
		\end{tabular}
	}
\end{table}

Almost all GDCUNet Settings have achieved competitive performance. Setting 5 achieves the highest scores on the primary overall metrics IoU and Dice with only 1.40M parameters. At the same time, there is also more lightweight option (Setting 3) that achieves a balance between parameter quantity and segmentation performance.
DconnNet has the highest recall but is limited overall.
Benefit by the characterization ability of SAFDConvolution for globally self-similar complex local features, GDCUNet Setting 5 has reached state-of-the-art comprehensive level.

\footnotetext[1]{Held in conjunction with MICCAI 2018.}
\footnotetext[2]{We emphasize conducting experiments under the unified benchmark and configuration. However, the model based on prior knowledge characterizing the vascular topological structure involves separate settings for data processing, loss functions, etc., and experiments need to be conducted under separate configurations.}

\begin{figure}[H]
	\raggedright
	\includegraphics[width=1.0\textwidth]{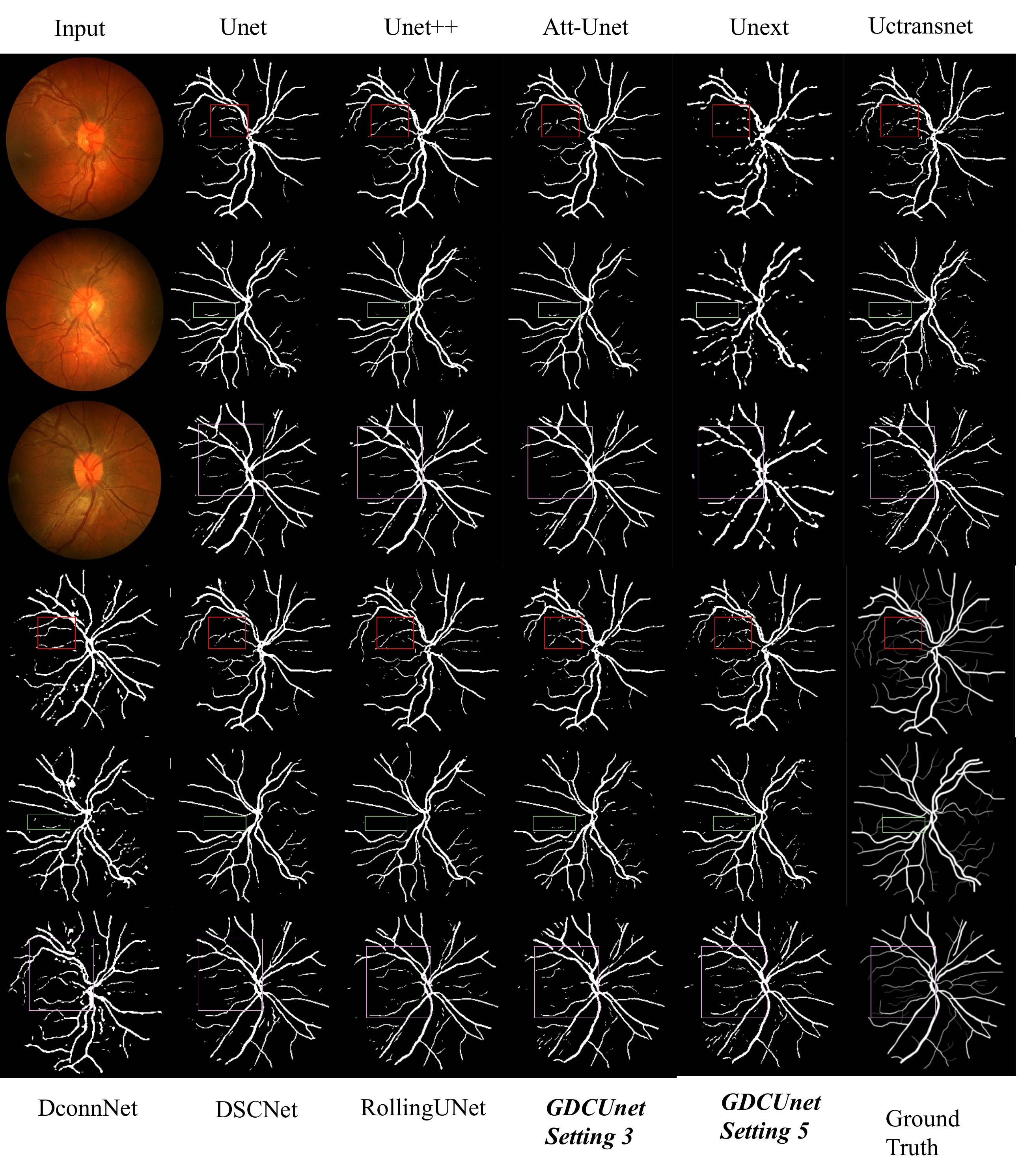}
	\caption{Visual comparison of segmentation results. Please zoom in for a better illustration.}
	\label{fig:results}
\end{figure}

Figure~\ref{fig:results} presents a qualitative comparison of the visual effects. DconnNet, which models the prior knowledge, has a relatively clear and complete segmentation mask, but more false positive examples have emerged. GDCUnet has a significantly better segmentation effect and can analyze the tiny blood vessels that are ignored by the exiting state-of-the-art models. Even the lightest Setting 3 can achieve competitive performance.

\section{Ablation Study}

\label{abl}

\subsection{The Effect of SAFDConvolution}

The main improvement made by GDCUNet is the addition of SAFDConvolution during the feature processing, thereby more effectively capturing complex local details with global self-similarity.
Compared with other existing deformable convolution methods, SAFDConvolution can handle global features more effectively. We replace the convolution in the SAFDConvolution Block with other convolution  (unify the kernel size \(5 \times 5\)), and the quantitative evaluation results are shown in Table~\ref{tab:abl}.

\begin{table}[H]
	\tiny
	\centering
	
	\caption{Comparison of convolution variants plugged into the SAFDConvolution Block.}
	\begin{tabular}{l c c }
		\toprule
		Conv Variant  & IoU  & Dice     \\
		\midrule 
		Conventional & 0.5833  & 0.7368  \\
			Deformable V1 & 0.6053   & 0.7472  \\
		Deformable V2	 & 0.5992   &   0.7438   \\
		Deformable V3 &  0.6130   & 0.7601    \\
	SAFDConvolution & \textcolor{green}{0.6304}  &  \textcolor{green}{0.7733}    \\
		
		\bottomrule
	\end{tabular}
	\label{tab:abl}
\end{table}

\begin{figure}[H]
	\centering
	\includegraphics[width=1.0\textwidth]{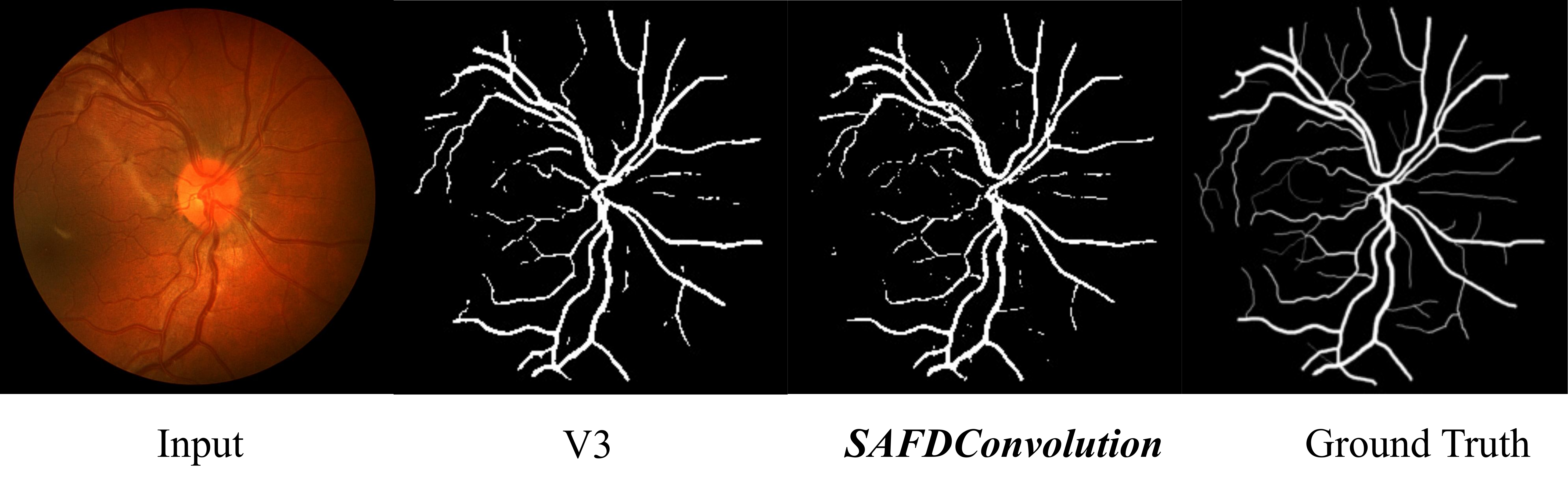}
	\caption{Comparison of convolution variants (Deformable V3 and our SAFDConvolution).}
	\label{fig:abl}
\end{figure}

Compared with the existing works, SAFDConvolution has significantly higher metrics. Visual comparison is shown in Figure~\ref{fig:abl}. Even compared with V3, the proposed module still has a significantly better visual effect\footnotemark[1].

\footnotetext[1]{The main focus of V4 is on the efficient computing of CUDA and the optimization of memory access. Its essential structure is similar to that of V3. For dynamic snake convolution, we have presented the results in DSCNet in Table \ref{tab:results}.}

\subsection{Feature Learning Significance}

\begin{figure}[H]
	\centering
	\includegraphics[width=1.0\textwidth]{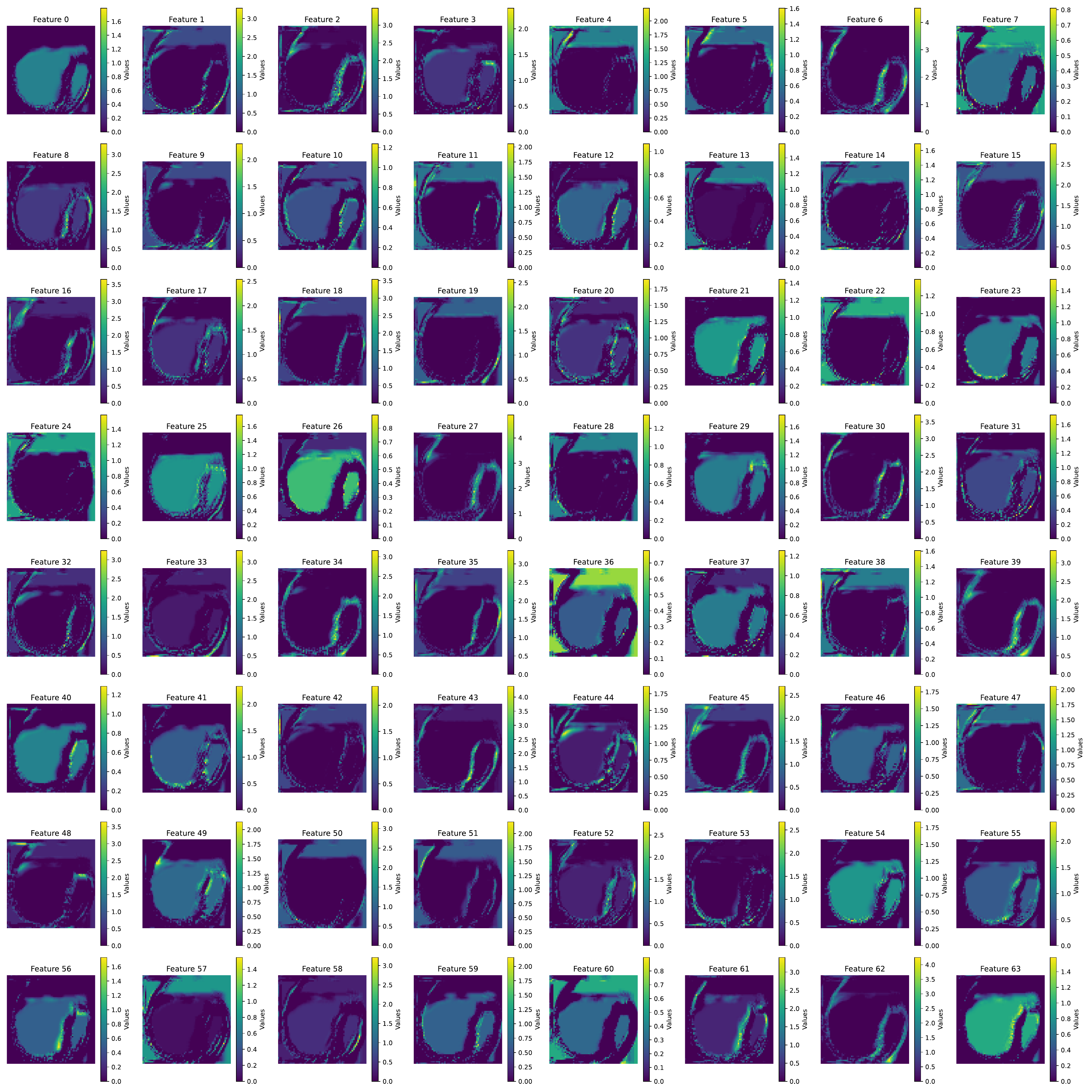}
	\caption{Visualization of feature maps obtained from SAFDConvolution. Please zoom in for a better illustration.}
	\label{fig:fea1}
\end{figure}

SAFDConvolution is sensitive to the complex local details of global self-similarity. To verify this, we present the feature maps obtained from SAFDConvolution during the encoding stage of GDCUnet in Figure~\ref{fig:fea1}.

\begin{figure}[H]
	\centering
	\includegraphics[width=1.0\textwidth]{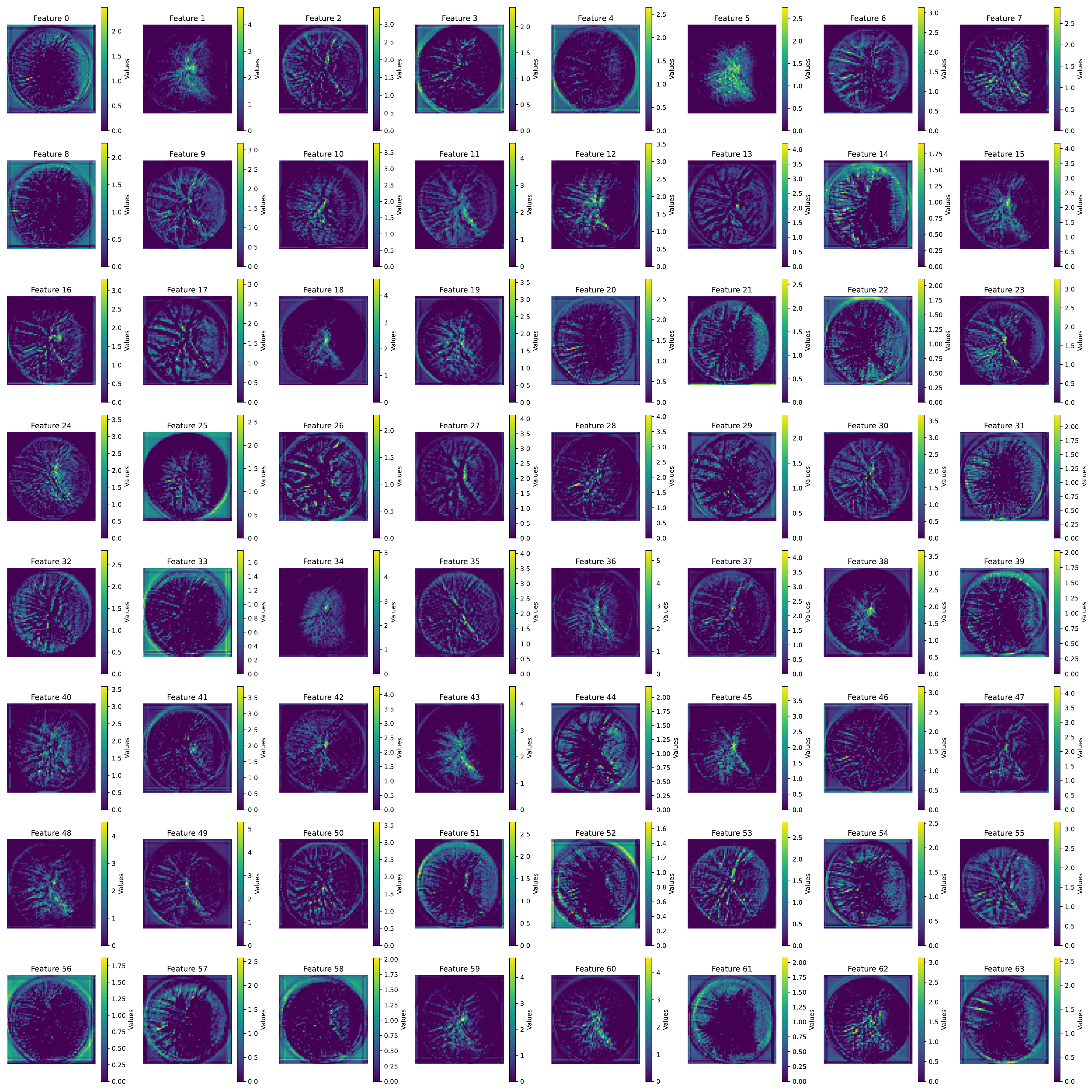}
	\caption{Visualization of feature maps obtained from conventional convolution. Please zoom in for a better illustration.}
	\label{fig:fea2}
\end{figure}

In other identical cases, only the convolutions plugged into SAFDConv Blocks are changed to conventional convolutions, the feature maps obtained at the same position are shown in Figure~\ref{fig:fea2}.

SAFDConvolution deforms the feature map with global features, thereby achieving relative deformation of the convolution kernel. Therefore, the resulting feature maps appear distorted to the human's eye. Compared with Figure~\ref{fig:fea2}, the feature maps in Figure~\ref{fig:fea1} show more highlighted features, indicating that the learning features are more significant.

\begin{figure}[H]
	\centering
	\includegraphics[width=1.0\textwidth]{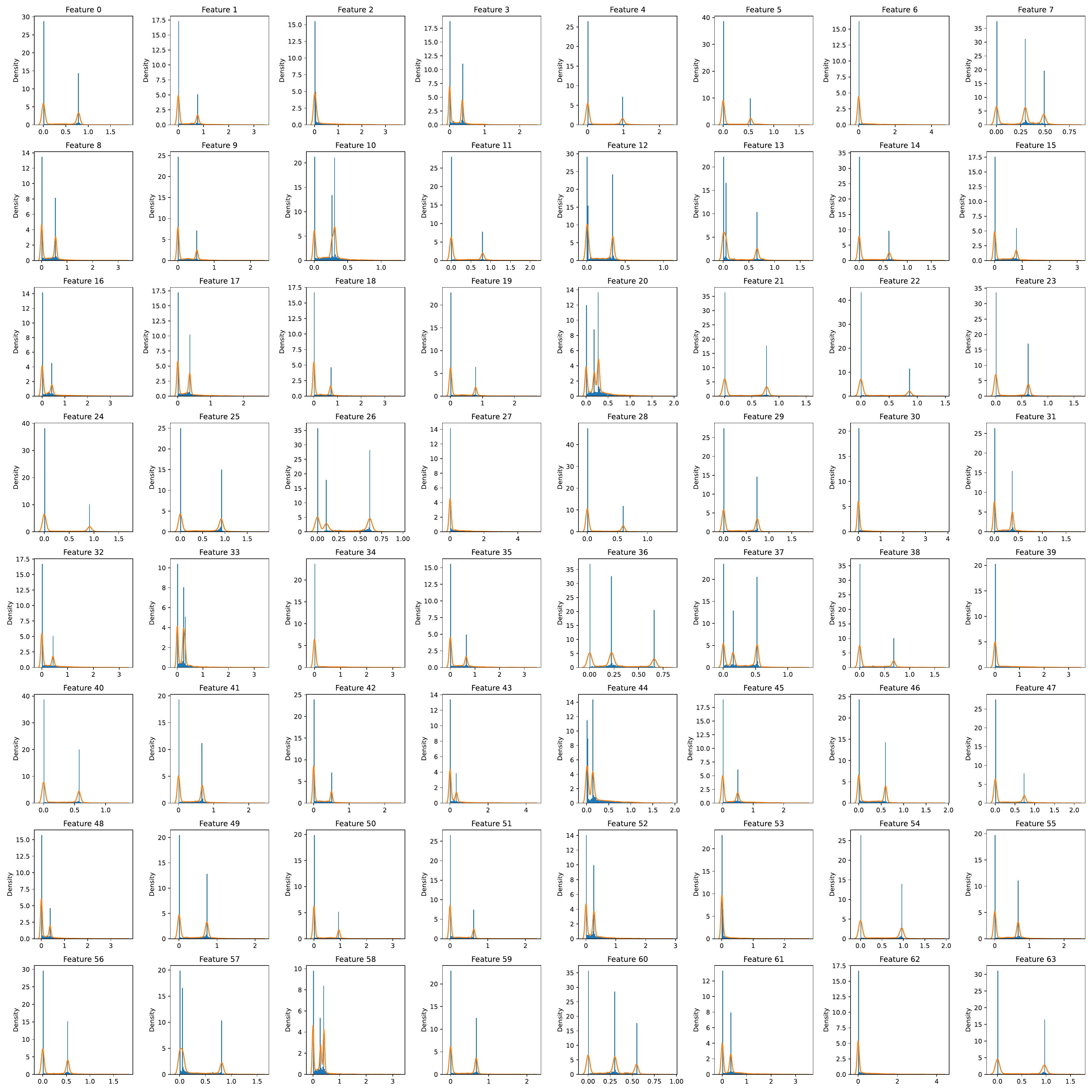}
	\caption{Distribution histograms of feature map grid values of SAFDConvolution. Please zoom in for a better illustration.}
	\label{fig:fea3}
\end{figure}

\begin{figure}[H]
	\centering
	\includegraphics[width=1.0\textwidth]{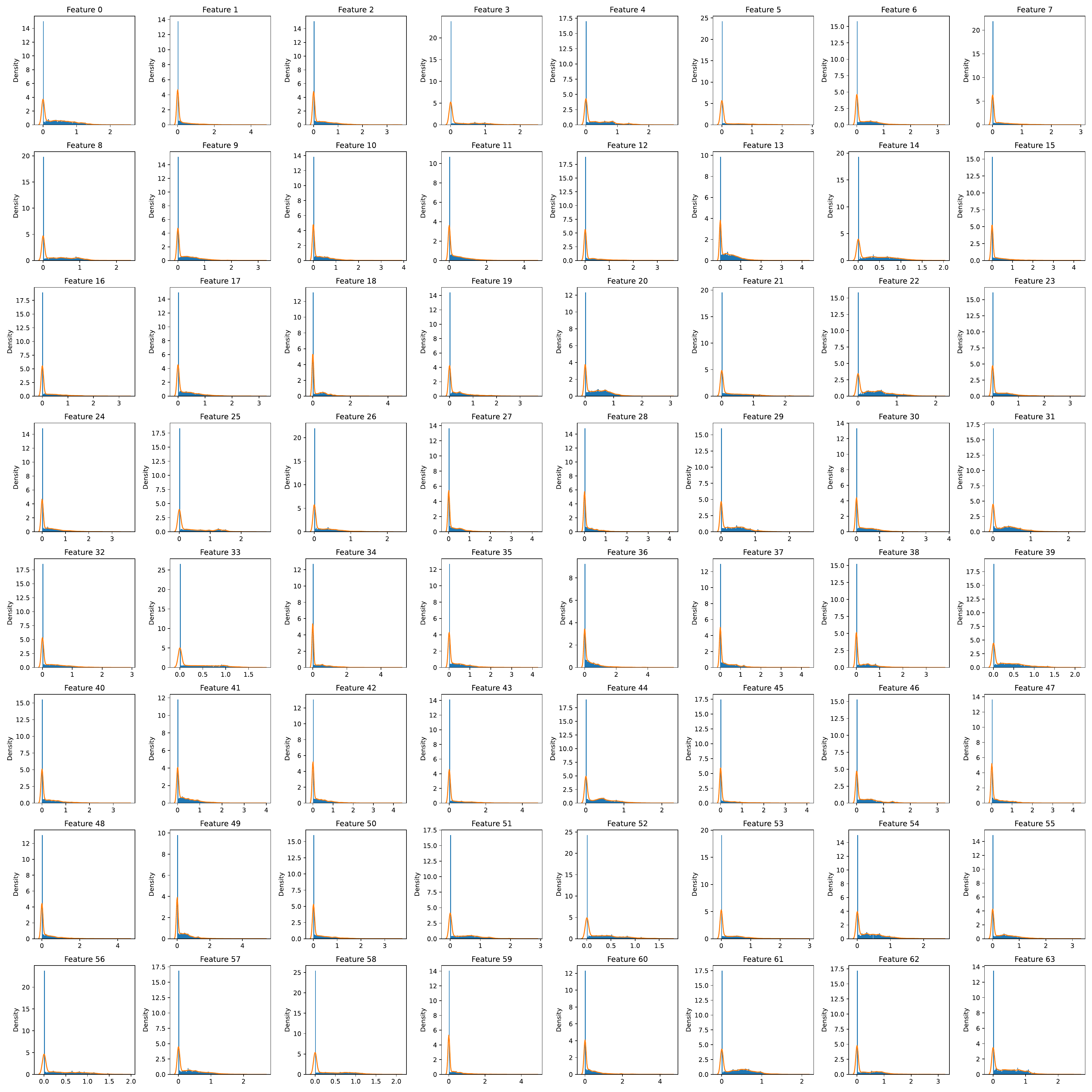}
	\caption{Distribution histograms of feature map grid values of conventional convolution. Please zoom in for a better illustration.}
	\label{fig:fea4}
\end{figure}

To further verify this point, we present the distributions of the feature map values. Figure~\ref{fig:fea3} corresponds to Figure~\ref{fig:fea1}, and Figure~\ref{fig:fea4} corresponds to Figure~\ref{fig:fea2}.

The distribution histograms of SAFDConvolution are relatively sharp, which indicates a high degree of feature separation, significant feature learning, and a strong target.

Compared with SAFDConvolution, the distribution of grid point values in feature maps obtained by conventional convolution under the same other conditions is smoother and more uniform. Conventional convolution kernels respond relatively evenly across the entire field of view, resulting in feature entanglement and a decline in discriminative ability.

Therefore, for vascular features that are locally complex and globally self-similar, SAFDConvolution can process the features more effectively and significantly.

\subsection{T-SNE Visualization}
To further verify the generalization performance of SAFDConvolution, following the approach of J. Feng et al. \cite{Tsne}, we use t-SNE to reduce the dimension of learned feature tensors. First, we visualize the training and testing features extracted by SAFDConv Block in GDCUnet. Then, we replace SAFDConvolutions with conventional convolutions while keeping other structures unchanged to visualize the training-testing features extracted by conventional convolutions. The results are shown in Figure 10, where the training features after dimensionality reduction are represented as orange dots and the testing features are represented as blue dots.

	\begin{figure}[H]
		\label{fig:tsnef}
	\centering
	\subfigure[]{
		\includegraphics[width=0.55\linewidth]{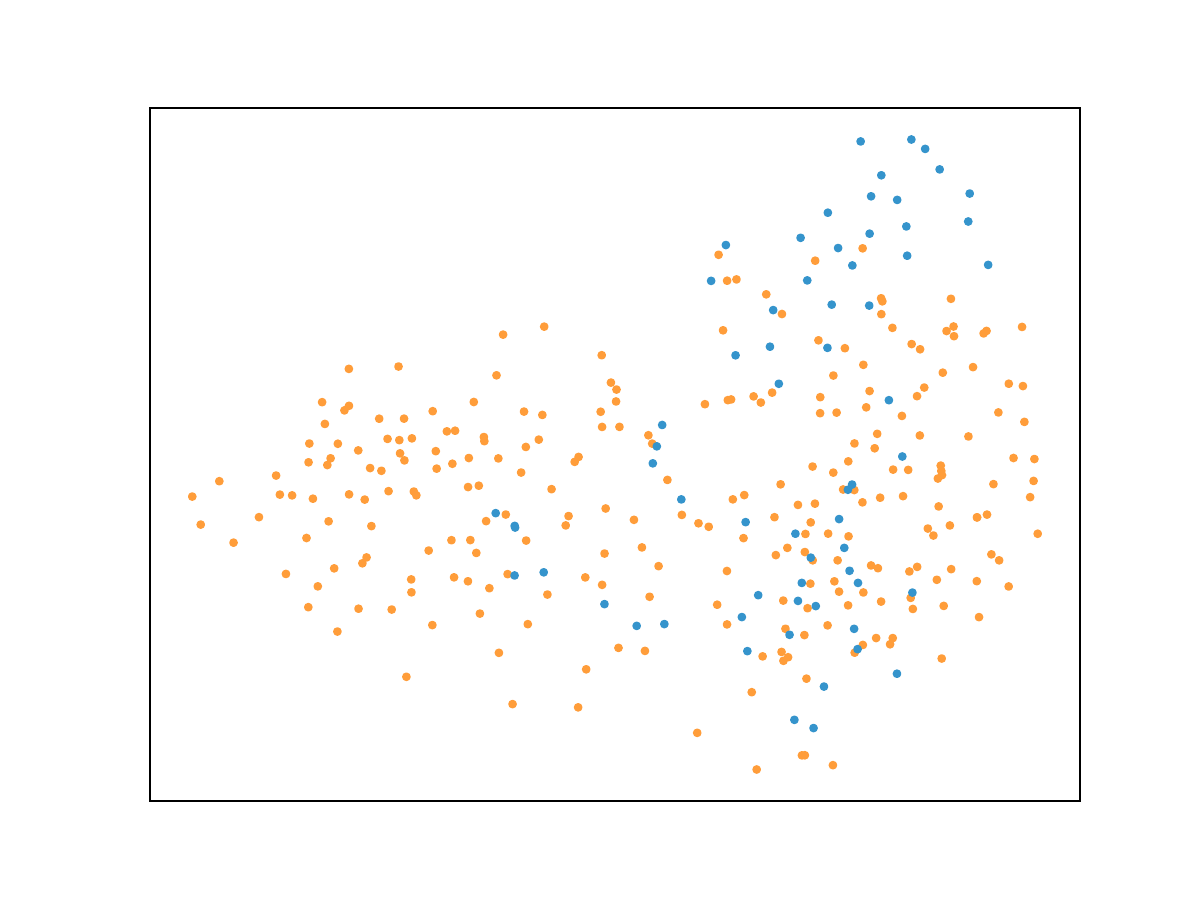}
	}
	\subfigure[]{
		\includegraphics[width=0.55\linewidth]{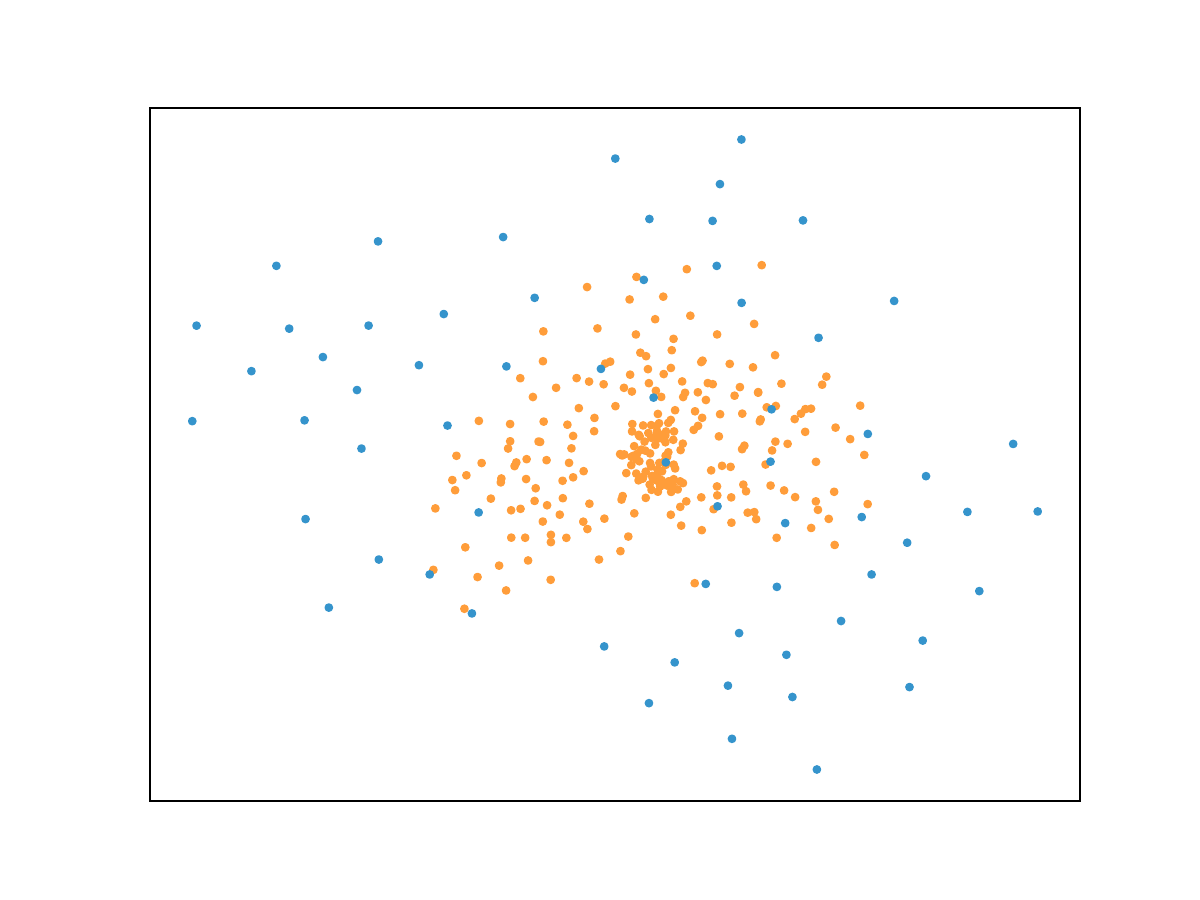}
	}
	
	\caption{T-SNE visualization of learned feature tensors. (a) SAFDConvolution. (b) Conventional convolution.}
\end{figure}

The feature distribution of conventional convolution in the testing set is relatively scattered, and the overlap between the training and testing domains is low, presenting a strong domain shift. The features learned by SAFDConvolution show a high degree of overlap between the training and testing domain samples within the same cluster, indicating that this convolution effectively extracts domain-invariant and highly discriminative representations, reduces domain shift, and has strong generalization ability.

\section{Conclusion}
\label{con}
To enhance the ability of deep learning models to represent locally complex but globally self-similar vascular features, we propose the plug-and-play SAFDConvolution, which is a novel type of deformable convolution.
Compared with existing deformable convolutions, SAFDConvolution achieves relative deformation and can support offset learning networks based on spatial attention.
SAFDConvolution is used to implement the proposed fundus vascular segmentation model GDCUNet. We establish a unified experimental benchmark. The experiments under this benchmark show that GDCUNet has reached state-of-the-art level. 
Ablation studies have verified that SAFDConvolution has stronger representational and generalization capabilities, can significantly learn key features, and notably improve the performance of the model.
Future work can focus on the application of SAFDConvolution in other machine vision tasks with similar features.

\textbf{Discussion of Limitations.} The introduction of spatial attention brings about computational overhead in matrix multiplication, which is a common problem for all Transformer-like structures and limits training and inference on the less powerful GPU. We hope to improve SAFDConvolution in the future by using methods such as sparse attention to further enhance its applicability.

\section*{CRediT authorship contribution statement}

\textbf{Lexuan Zhu}: Data curation, Formal analysis, Investigation, Methodology, Resources, Software, Visualization, Writing - original draft.
\textbf{Yuxuan Li}: Resources, Software, Validation, Writing - original draft.
\textbf{Yuning Ren}: Methodology, Conceptualization, Funding acquisition, Investigation, Supervision, Validation, Writing - review and editing.

\section*{Declaration of interests}
The authors declare that they have no known competing financial interests or personal relationships that could have appeared to influence the work reported in this paper.
\section*{Data availability}
The dataset used in this research is publicly available. All codes are available in our project website, mentioned in abstract. 
\section*{Acknowledgments}
This work was supported by the Ministry of Education of the People’s Republic of China under Grant 2024GH-ZDA-GJ-Y-09.

\end{document}